\documentclass[10pt,twocolumn,letterpaper]{article}

\usepackage{times}
\usepackage{epsfig}
\usepackage{graphicx}
\usepackage{amsmath}
\usepackage{amssymb}
\usepackage{mwe} 
\usepackage{subcaption}
\usepackage{array}
\usepackage{booktabs}
\usepackage{float}

\newcommand{\bftab}{\fontseries{b}\selectfont}

\usepackage[breaklinks=true,bookmarks=false]{hyperref}

\begin{document}

%%%%%%%%% TITLE
\title{Refinement of Predicted Missing Parts Enhance Point Cloud Completion}
\date{}

\author{Alexis Mendoza\\
Universidad Nacional de San Agust\'{i}n\\
Arequipa, Per\'{u}\\
{\tt\small amendozavil@unsa.edu.pe}
\and
Alexander Apaza\\
Universidad Nacional de San Agust\'{i}n\\
Arequipa, Per\'{u}\\
{\tt\small aapazato@unsa.edu.pe}
\and
Ivan Sipiran\\
Department of Computer Science, University of Chile\\
Santiago, Chile\\
{\tt\small isipiran@dcc.uchile.cl}
\and
Cristian L\'{o}pez\\
Universidad La Salle\\
Arequipa, Per\'{u}\\
{\tt\small clopez@ulasalle.edu.pe}
}

\maketitle

\begin{abstract}
   Point cloud completion is the task of predicting complete geometry from partial observations using a point set representation for a 3D shape. Previous approaches propose neural networks to directly estimate the whole point cloud through encoder-decoder models fed by the incomplete point set. By predicting the complete model, the current methods compute redundant information because the output also contains the known incomplete input geometry. This paper proposes an end-to-end neural network architecture that focuses on computing the missing geometry and merging the known input and the predicted point cloud. Our method is composed of two neural networks: the missing part prediction network and the merging-refinement network. The first module focuses on extracting information from the incomplete input to infer the missing geometry. The second module merges both point clouds and improves the distribution of the points. Our experiments on ShapeNet dataset show that our method outperforms the state-of-the-art methods in point cloud completion.  The code of our methods and experiments is available in \url{https://github.com/ivansipiran/Refinement-Point-Cloud-Completion}.
\end{abstract}

\section{Introduction}
\label{Sec:Introduction}

Point cloud analysis is among the most fruitful topics in the deep learning era. Pioneering techniques such as PointNet~\cite{sa:Qi2017a} and PointNet++~\cite{sa:Qi2017b} have opened a myriad of possibilities for the development of useful analysis tools for point clouds such as classification, segmentation, and reconstruction, to name a few. The preference of a point cloud representation over other representations such as views, voxels, or meshes is mainly due to its compactness and versatility. Also, point clouds are the de-facto representation produced by many devices such as laser scanners, RGB-D scanners, or stereo vision cameras.

\begin{figure}[htb]
\begin{minipage}[t]{.47\textwidth}
\centering
\includegraphics[width=\textwidth]{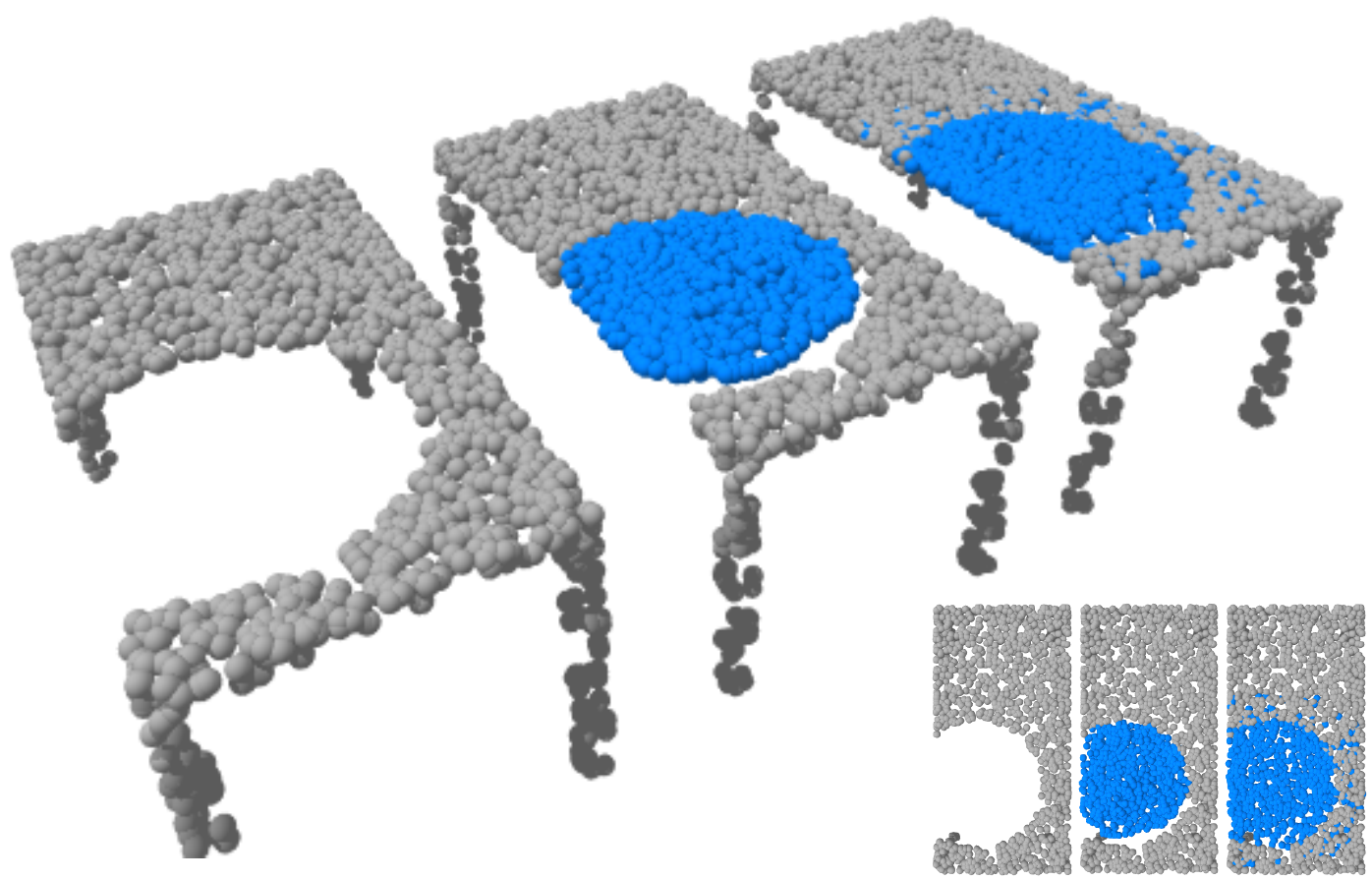}
\end{minipage}
\caption{Our solution takes an incomplete point cloud as input (left). A Missing Part Prediction Network infers the missing part and concatenate it with the input (center). Finally, a Point Refiner Network improves the distribution of points and the geometric transition between the merged sets (right).}
\label{fig:principal}
\end{figure}

The typical strategy to address the problem of point cloud completion is the encoder-decoder model. In this model, the encoder transforms the incomplete point cloud into an intermediate representation that conveys the input's main geometric characteristics. Subsequently, the decoder converts the intermediate representation into the completed point cloud. Typically, this approach requires a refinement step to ensure some output properties, such as level of resolution or point distribution. Nevertheless, previous methods try to reconstruct the complete point cloud from the input, which may lead to undesired results such as the loss of the original point cloud distribution.

An ideal completion algorithm should hold one crucial property: to keep the original incomplete data unchanged as much as possible. This property is particularly important in real-world scenarios, where the data comes from sensing devices, and we could be interested in keeping the main characteristics of the incoming geometry. Precisely, our proposal addresses the point cloud completion problem taking into account the property above. We propose an end-to-end neural network that focuses on predicting the missing part from a partial observation of a given object. By predicting only the missing geometry, the input point cloud remains unchanged, and therefore we can leverage its structure to produce a final reconstruction. Our proposal still requires a refinement step to improve the overall distribution of the resulting point cloud and to close the gaps between the missing part and the original input (see Fig.~\ref{fig:principal}). Nevertheless, our results show that preserving the original incomplete point cloud leads to superior results concerning previous approaches.

The contributions of our paper can be summarized as follows:
\begin{itemize}
\item A method that focuses on predicting the missing part of the analyzed point cloud with a learning process controlled by a specific loss function.
\item A refinement strategy to guarantee a good distribution of the final point cloud.
\item A simple and effective overall architecture for point cloud completion.
\item A set of experiments to show the effectiveness and robustness of our proposal. 
\end{itemize}

Our paper is organized as follows. Section~\ref{Sec:Related} shows the related works. Section~\ref{Sec:Method} describes our architecture and the loss function. Section~\ref{Sec:Experiments} presents our experiments on the ShapeNet dataset to perform the completion in point clouds. Finally, Section~\ref{Sec:Conclusions} concludes our paper and gives ideas for future work.

\section{Related Work}
\label{Sec:Related}
Several techniques have been proposed to address the problem of shape completion. Most of these techniques tackled shape completion assuming geometric priors such as symmetries or content-based matching. Sipiran et al.~\cite{sa:Sipiran2014} proposed an algorithm for finding symmetric correspondences in 3D shapes, and subsequently, these correspondences were used to predict missing geometry. Similarly, Sung et al.~\cite{sa:Sung2015} used symmetry priors to improve the search of a template shape in a database, subsequently used to complete the input shape. Harary et al.~\cite{sa:Harary2014} devised an algorithm to recover missing geometry using the self-similarity of local patches in shape. Nevertheless, the geometric priors impose a strong assumption for the completion task, making this strategy not always feasible.

Due to the rise of deep learning techniques and the availability of large-scale 3D benchmarks, many efforts have been made in devising shape completion algorithms in a data-driven manner. The first proposals took advantage of the progress of convolutional neural networks(CNN) in the computer vision field to manage volumetric representations. Dai et al.~\cite{sa:Dai2017} proposed a 3D encoder predictor network that completes an input scan. Similarly, Han et al.~\cite{sa:Han2017} used an LSTM network to provide a high-resolution volumetric completion. The same progress in CNN networks also enabled the image-based representation of 3D shapes. Hu et al.~\cite{sa:Hu2019a} developed a generative adversarial network(GAN) that completes rendered depth-maps from incomplete point clouds. The completed depth-maps are thus reprojected to reconstruct the original 3D incomplete shape. Also, Hu et al.~\cite{sa:Hu2020} improved the way to evaluate the loss of the depth-maps generated by a GAN to introduce geometric consistency.

The success of the point cloud representation for a 3D shape is mainly due to the advent of neural networks that can directly process unordered sets. The base for all the current progress in point cloud analysis is the PointNet architecture~\cite{sa:Qi2017a} and its hierarchical variant PointNet++~\cite{sa:Qi2017b}. A significant result of these architectures is the possibility of computing a high-level feature vector that conveys the geometric information of a point cloud in a condensed manner. Also, Achlioptas et al~\cite{sa:Achlioptas2018} motivated the use of generative models for point clouds, showing the representational power of encoder-decoder architectures for this kind of data. Yuan et al.~\cite{sa:Yuan2018} developed a completion algorithm in two stages to cope with coarse and fine point cloud decoding. Similarly, Tchapmi et al.~\cite{sa:Tchapmi2019} proposed a tree-based decoder to give the completion task the multi-resolution capacity. Also, Liu et al.~\cite{sa:Liu2020} presented a two-stage completion algorithm by mixing a patch-wise folding decoder and a density sampling merging to deal with good coverage of the object surface.

\begin{figure*}[ht]
\centering
\includegraphics[width=\textwidth]{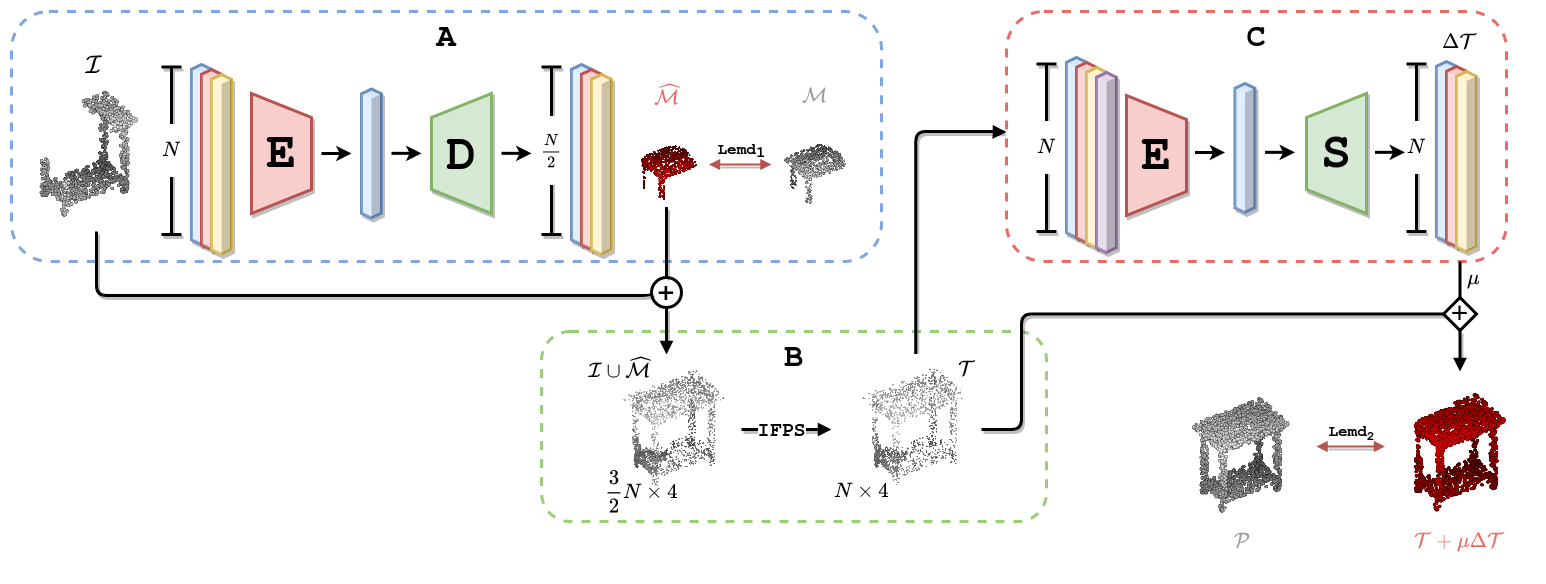}
\caption{Workflow of our method. It consists of three main stages. First, the Missing Part Prediction Network (Section \ref{pcn}, module A) takes an incomplete point cloud $\mathcal{I}$ and outputs the corresponding missing part $\widehat{\mathcal{M}}$. Second, the Merging and Sampling strategy concatenate and sample the incomplete input cloud and the predicted missing part (Section \ref{sampling}, module B). Third, the Point Refiner Network improves the distribution of the point cloud and closes the gaps generated by the merging (Section \ref{prn}, module C).}
\label{fig:main_arquitecture}
\end{figure*}

More recently, we witnessed the rise of more elaborated approaches to tackle point cloud completion. Most of these approaches use modern strategies in deep learning, trying to surpass the problem of generating visually appealing shapes. For instance, Wang et al.~\cite{sa:Wang2020} proposed an algorithm to generate a coarse point cloud subsequently refined through a lifting module that upsamples the generated point cloud. Likewise, Huang et al.~\cite{sa:Huang2020} preferred to predict the missing region with a multi-resolution point pyramid decoder, inspired by the feature pyramid network in computer vision tasks. On the other hand, the idea of the attention mechanism to guide the data generation has been recently applied to point cloud completion. For instance, Sun et al.~\cite{sa:Sun2020} developed a conditional generative model that includes an attention mechanism to weight point features according to their importance in the reconstruction of the point cloud. Similarly, Wen et al.~\cite{sa:Wen2020} devised an encoder-decoder architecture with an attention mechanism and hierarchical folding-based decoder to complete point clouds.

\section{Method}
\label{Sec:Method}
Let $\mathcal{P}$ be the point cloud representing a 3D shape. We assume that $\mathcal{P}$ is split in the incomplete point cloud $\mathcal{I}$ and the missing part $\mathcal{M}$ such that $\mathcal{P} = \mathcal{I} \cup \mathcal{M}$. Figure~\ref{fig:main_arquitecture} shows our proposed workflow to reconstruct $\mathcal{P}$ from the observed point cloud $\mathcal{I}$. Our method consists of three modules that operate in an end-to-end manner to complete a given input.

The first module (block A in Figure~\ref{fig:main_arquitecture}) is a learned parametric function that computes the missing part $\widehat{\mathcal{M}}$. During training, our method controls the learning process by applying a loss function to measure the similarity between the predicted missing part $\widehat{\mathcal{M}}$ and the real missing part $\mathcal{M}$. Recently, the Point Fractal Network(PF-Net)~\cite{sa:Huang2020} also proposed to generate the missing part from the incomplete input. The differences between our method and PF-Net are two-fold: a) our architecture is simple, and b) our explicit loss function for the missing part helps to guide the end-to-end learning. Our experiments confirm that the design introduced by our method outperforms the performance of PF-Net.

The second module (block B in Figure~\ref{fig:main_arquitecture}) performs the union of the incomplete input $\mathcal{I}$ and the predicted missing part $\widehat{\mathcal{M}}$. The simple concatenation of these two point sets could lead to problems in the point distribution of the object and the transition between both point sets. Therefore, this module is responsible for improving the sampling of the point set $\mathcal{I} \cup \widehat{\mathcal{M}}$ using the recently proposed Minimum Density Sampling technique~\cite{sa:Liu2020}. The result of this module is a sampled point set $\mathcal{T}$.

The third module (block C in Figure~\ref{fig:main_arquitecture}) is a learned parametric function that refines the sampled point set $\mathcal{T}$. This function computes a displacement field $\Delta\mathcal{T}$ that shifts the points of $\mathcal{T}$ to the desired position to improve the point distribution. The final output of our proposal is a point cloud $\mathcal{T} + \mu\Delta\mathcal{T}$, where $\mu$ is a hyperparameter that controls the amount of displacement we want to keep. During training, the learning process of this module is controlled by a loss function that measures the similarity between the predicted point cloud $\mathcal{T} + \mu\Delta\mathcal{T}$ and the ground-truth $\mathcal{P}$.

Section~\ref{pcn} introduces the architecture of our neural network for the missing part prediction. Section~\ref{sampling} describes the merging and sampling method. Section~\ref{prn} presents the architecture of our refinement network. Finally, Section~\ref{loss} defines the joint loss function for training our model.

\subsection{Missing Part Prediction Network (MPN)} \label{pcn}

The main goal of this network is to predict the missing part from an incomplete point cloud. This task is performed by extracting features from the incomplete input with an encoder network. For efficiency, we adopt a PointNet \cite{sa:Qi2017a} based encoder; nevertheless, any feature extractor network could be used. We then use a decoder network to obtain the missing part from the feature vector. For the decoder, we use a Morphing-based Decoder as proposed by Liu et al.~\cite{sa:Liu2020}. In addition, we also test a Multi-layer Perceptron decoder to see the impact of the decoding in the missing part prediction. In our experiments, the model that uses the morphing-based decoder is named \emph{Ours-MBD}, and the model with the MLP decoder is named \emph{Ours-MLP}.

\begin{figure*}
\centering
\includegraphics[scale=0.3]{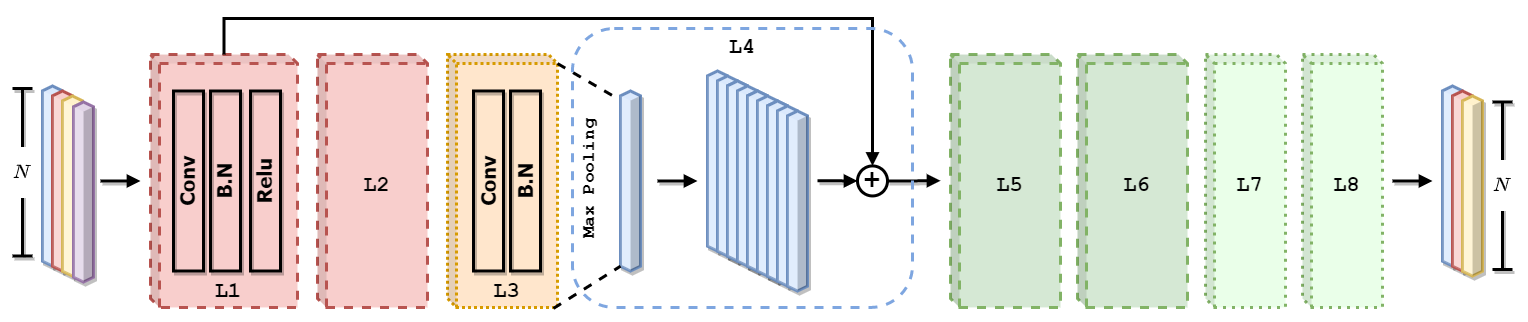}
\caption{Point Refiner Network. It takes a labeled point cloud (the purple column represents the label) and passes it through a PointNet Residual Network (L1 expands the point cloud to 64, L2 to 128, and L3 to 1024) to generate a global feature for the point cloud. Then we replicate it 1024 times and concatenate it with the result of L1. Layers L5, L6, L7, and L8 reduce the concatenation and output a three-channel displacement field.}
\label{fig:residual_network}
\end{figure*}

\subsubsection{Missing Part Network Encoder}
We adopt a PointNet based encoder that obtains a feature vector $FV$ from a $N\times3$ incomplete point cloud. The encoder architecture consists of four block layers. The first (L1) and second (L2) blocks consist of three layers: 1D-convolution, batch-normalization, and ReLU activation function. In contrast, the third block (L3) has only a 1D-convolution layer and a batch-normalization layer. The number of kernels for 1D convolutions in L1, L2, and L3 are 64, 128, and 1024, respectively. Also, the kernels have a size of one, a stride of one, and padding of zero. Finally, the fourth layer is a fully-connected layer with 1024 neurons. The result is a global feature vector of dimension 1024.

\subsubsection{Missing Part Network Decoder}
The decoder must transform the feature vector $FV$ into the desired missing part. We denote the output missing part as a point cloud of size $M \times 3$. To compare the overall architecture performance, we propose to use two different networks for the decoder: A Multi-Layer Perceptron (MLP) and a Morphing Based Decoder (MBD)\cite{sa:Liu2020}.

The MLP decoder consists of three building blocks. Each block $L_i$ consists of fully connected layers followed by a ReLU activation with the exception of the third block. The fully connected layers have a number of neurons of $1024,\ 1024,\ M \times 3$). For the Morphing-based Decoder, we use the architecture described in \cite{sa:Liu2020}. This decoder consists of $K$ (16 in experiments) morphing networks. Each network maps a 2D surface, sampled from the unit square $[0, 1]^2$, to a 3D surface. In each forward, we sample $(M/k)$ points in the unit square. The feature vector $FV$ is then concatenated to each point coordinate, before passing through the $K$ morphing networks. Each sampled 2D points are then mapped to a 3D surface. Resulting in $M$ points describing the predicted shape.

Each morphing network has four blocks. The first three blocks consist of a convolution, followed by batch normalization and a ReLU activation function. The last block has a convolution layer and a $tanh$ activation function.

\subsection{Merging and Sampling} \label{sampling}
The input (incomplete) point cloud $\mathcal{I}$ is concatenated to the predicted missing part $\widehat{\mathcal{M}}$. The result of the concatenation is a point cloud of size $\frac{3}{2}\times N \times 3$. We realize that the concatenation has two issues. First, the density of the missing part $\widehat{\mathcal{M}}$ is higher than the density of the input point clouds $\mathcal{I}$. Second, the size of our concatenated point cloud does not match the size of the ground truth. To address these problems, we use a sampling method to adjust point distribution and the size of the merged point cloud. We use the iterative farthest point sampling (IFPS), which has been effectively applied in PointNet++ \cite{sa:Qi2017b}.

\subsection{Point Refiner Network (PRN)} \label{prn}
The concatenation of the incomplete input $\mathcal{I}$ and the predicted missing part $\widehat{\mathcal{M}}$ may produce a visible crack in the resulting point cloud. The goal of our refiner network is to solve this issue by improving the transition between the merged point clouds. Besides, the refiner network also can enhance the final distribution of the points in the object. The architecture of the Point Refiner Network(PRN) is in Figure \ref{fig:residual_network}. The input to the PRN module is the sampled point cloud of size $N \times 3$.
Nevertheless, we augment the fourth channel to the input point cloud to assign a label for each point. We assign a zero label to points belonging to the original incomplete point cloud $\mathcal{I}$, and one label to points belonging to the predicted missing part $\widehat{\mathcal{M}}$. This augmented channel is used as extra information by the refiner network to decide which points require changes to improve the reconstruction.

The Point Refiner Network has eight layers. Layers L1, L2, and L3 consist of 1D-convolutions, followed by a batch normalization layer. A ReLU activation function also follows L1 and L2. The number of kernels of the first three convolutional layers is 64, 128, and 1024, respectively. Layer L4 applies a max-pooling operator to the output of L3 and replicates the output $N$ times. The resulting tensor of size $(1024 \times N)$ is then concatenated to the output of L1, resulting in a tensor of size ($1088 \times N$), which is then passed through the next 1D-convolutions layers: L5, L6, L7, and L8. These layers are followed by batch normalization and a ReLU activation function, except the last layer L8, which has a Tanh activation function. The number of kernels of the convolutional layers L5, L6, L7, and L8, are 512, 256, 128, and 3, respectively.

The Point Refiner Network is intended to predict a displacement field $\Delta\mathcal{T}$. The final output of the whole network is the point cloud $\mathcal{T} + \Delta\mathcal{T}$. Figure~\ref{fig:principal} shows the effect of the refinement in our proposal.

\subsection{Loss Function}
\label{loss}

One crucial step for the reconstruction is to define a proper reconstruction loss. In point cloud analysis, there exist predominantly two loss functions:
Earth Mover's Distance (EMD) and Chamfer Distance (CD). Given two point clouds $S_1$ and $S_2$, CD measures the mean distance between each point in $S_1$ to its spatial nearest neighbor in $S_2$ plus the mean distance between each point in $S_2$ to its spatial nearest neighbor in $S_1$.

In contrast, EMD is a metric between two distributions based on the minimal cost of transforming one distribution into the other. Given two point clouds of the same size, EMD is defined as follows:

\begin{equation}
L_{EMD}(S_1, S_2) = \underset{\phi : S_1 \rightarrow S_2 }{min} \frac{1}{|S_1|} \sum_{x \in S_1} {\|x - \phi(x)\|}_2
\end{equation}

\noindent where $\phi$ is a bijection. Many works use the Chamfer distance because it is less computationally intensive than EMD. Nevertheless, EMD provides better reconstruction results due to its one-to-one point mapping. In this work, we used the EMD implementation presented in \cite{sa:Liu2020}, which has a $O(n)$ memory footprint.

\begin{table*}[t]
\centering
\selectfont\begin{tabular}{llllll}
\hline
{\bftab Categories} & {\bftab MSN} \cite{sa:Liu2020} & {\bftab PF-Net} \cite{sa:Huang2020} & {\bftab FCAE} & {\bftab Ours-MBD} & {\bftab Ours-MLP} \\ \hline
Airplane & 3.113 / 3.500 & 4.604 / 2.997 & 6.980 / 7.047 & {\bftab 2.147 } / 2.528 & 2.892 / {\bftab 2.212} \\
Bag & 15.393 / 7.942 & 21.350 / 7.184 & 34.774 / 29.045 & {\bftab 9.975 } / 6.731 & 14.232 / \textbf{5.748} \\
Cap & 16.824 / 5.842 & 28.984 / 3.719 & 52.914 / 37.094 & {\bftab 8.596 } / {\bftab 2.500 } & 17.016 / 2.527 \\
Car & 8.443 / 6.991 & 9.643 / 3.206 & 19.982 / 15.550 & {\bftab 6.539 } / 2.927 & 8.683 / \textbf{2.516} \\
Chair & 5.934 / 4.434 & 7.803 / 2.774 & 16.408 / 13.733 & \textbf{4.371} / 2.259 & 6.049 / \textbf{2.051} \\
Guitar & 5.226 / 3.365 & 3.195 / 3.068 & 3.276 / 3.185 & {\bftab 1.329 } / 2.483 & 1.822 / \textbf{2.381} \\
Lamp & 19.873 / 9.819 & 24.143 / 10.559 & 39.082 / 21.319 & {\bftab 12.712 } / 10.545 & 15.675 / \textbf{9.389} \\
Laptop & 3.543 / 3.910 & 4.879 / 2.262 & 9.252 / 9.623 & {\bftab 2.239 } / 1.551 & 2.832 / {\bftab 1.430} \\
Motorbike & 6.802 / 7.178 & 6.473 / 5.028 & 14.094 / 11.592 & {\bftab 4.897 } / 5.946 & 6.197 / \textbf{4.162} \\
Mug & 8.199 / 6.340 & 10.538 / 3.774 & 25.223 / 22.119 & {\bftab 4.467 } / {\bftab 2.963 } & 7.464 / 3.054 \\
Pistol & 5.092 / 6.349 & 7.613 / 4.944 & 10.898 / 10.115 & {\bftab 4.137 } / 4.539 & 5.399 / \textbf{3.952} \\
Skateboard & 3.441 / 3.766 & 5.562 / 2.788 & 8.981 / 8.996 & {\bftab 2.250 } / 2.295 & 2.693 / \textbf{1.994} \\
Table & 7.257 / 4.709 & 9.919 / 3.454 & 19.712 / 17.699 & {\bftab 5.556 } / 2.777 & 6.722 / \textbf{2.737} \\ \hline
{\bftab Average} & 8.395 / 5.704 & 11.131 / 4.289 & 20.121 / 15.932 & {\bftab 5.324 } / 3.849 & 7.514 / \textbf{3.396 }
\\ \hline
\end{tabular}
\caption{Quantitative comparison on ShapeNet dataset using average squared distance \cite{sa:Huang2020}. Values are shown in pairs where the first value is $Pred \rightarrow GT$ error and the second value is $GT \rightarrow Pred$ error. Results are scaled by $10,000$.}
\label{tab:comparisonShapeNet}
\end{table*}

In the proposed architecture, we have two losses calculated for both the \textbf{MPN} and the \textbf{PRN} networks. The first loss calculates the EMD between the predicted missing point cloud $\widehat{\mathcal{M}}$ and the missing ground truth point cloud $\mathcal{M}$. The second loss compares the refined point cloud $\widehat{\mathcal{P}}$ with the complete ground truth point cloud $\mathcal{P}$. Our joint loss is calculated as $L = L_\text{EMD}(\widehat{\mathcal{M}},\ \mathcal{M})\ +\ L_\text{EMD}(\widehat{\mathcal{P}},\ \mathcal{P}).$

\section{Experiments}
\label{Sec:Experiments}

\subsection{Data Generation and Training}
We evaluate our proposal using the same protocol defined in ~\cite{sa:Huang2020}. The dataset is composed of 13 classes from the ShapeNet-Part dataset~\cite{sa:Yi2016}: airplane, bag, cap, car, chair, guitar, lamp, laptop, motorbike, mug, pistol, skateboard, and table. The dataset contains 14473 models (11705 for training and 2768 for testing). Before generating the point clouds from the models, we normalize the input shapes' position and scale. To normalize the position, we translate the objects so that the object's centroid is in origin. To normalize the scale, we inscribe the input object inside a sphere of radius one. We finally sample 8192 points from the surface for each CAD model.

For each epoch, we generate a partial point cloud for each model. To produce a partial point cloud, we randomly choose the model's point and a radius of $r=0.35$. Subsequently, we split the complete point cloud into two sets: the points outside the sphere (partial point cloud) and the points within the sphere (missing part ground-truth). Finally, the complete point cloud is sampled to 2048 points, the partial point cloud is sampled to 2048 points, and the missing part is sampled to 1024 points.

The implementation of our models is in PyTorch. We use the ADAM optimizer with a learning rate of $0.001$. We use the $O(n)$ variant of the Earth Mover Distance as loss function as proposed in \cite{sa:Liu2020}. The models were trained in an NVIDIA GeForce RTX 2080 GPU for 200 epochs with a batch size of 64.

\begin{figure*}
\centering
\includegraphics[width=0.9\textwidth]{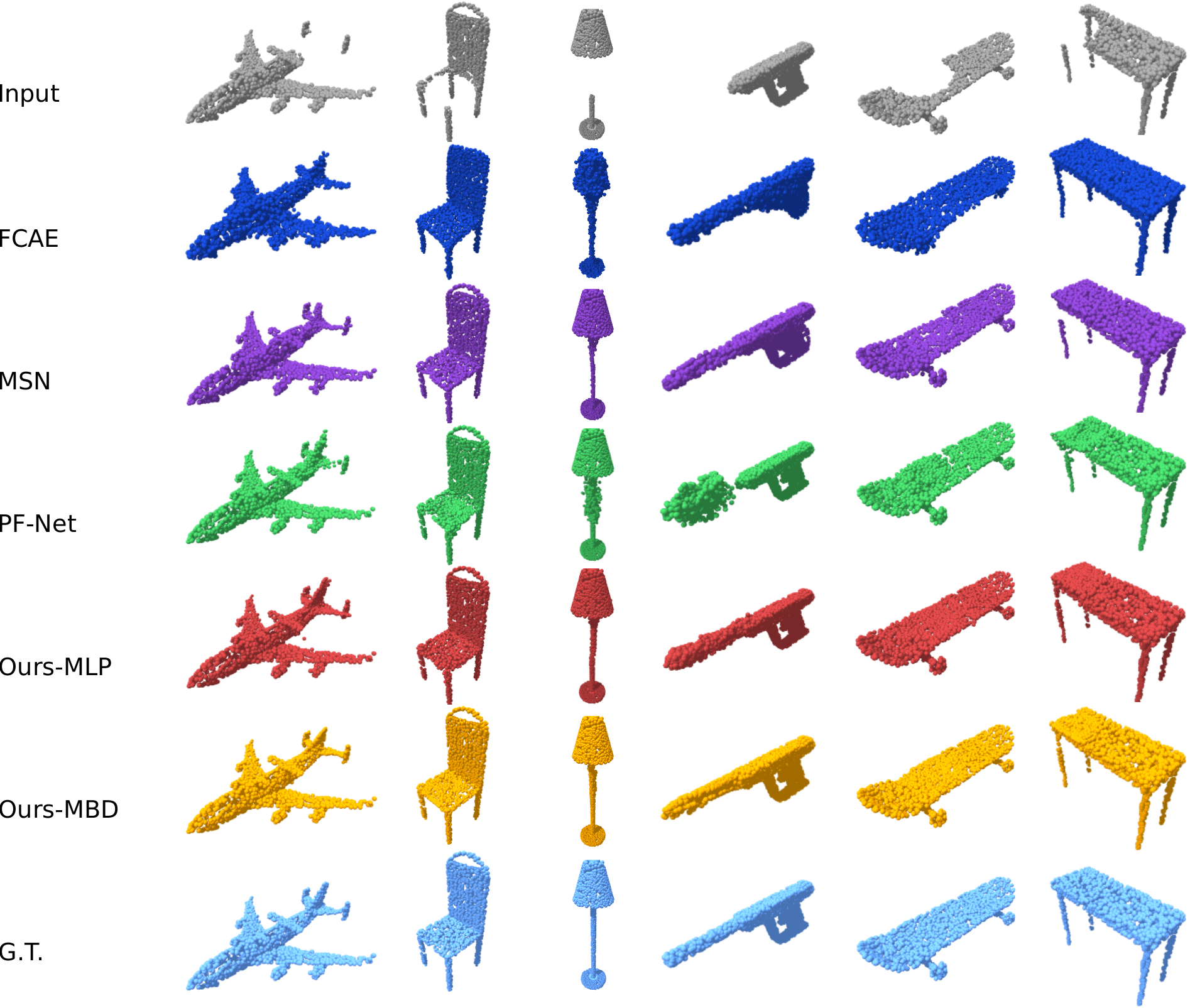}
\caption{Comparison of completion results between our method and state-of-the-art methods. FCAE predicts reasonable coarse shapes but fails to complete details. MSN~\cite{sa:Liu2020} produces the entire shape but struggles to find some details. PF-Net~\cite{sa:Huang2020} predicts the coarse missing part, but it cannot incorporate the prediction with the input. Our methods (Ours-MBD and Ours-MLP) successfully predict the missing part and integrate it to produce a good result.}
\label{fig:comparison}
\end{figure*}

\subsection{Comparison with Previous Methods}
\label{Sec:Comparison}
We compare our methods with state-of-the-art techniques and present quantitative and qualitative comparisons on the ShapeNet-Part test data. In this comparison \emph{Ours-MBD} denotes the method that uses a morphing-based decoder for the missing part prediction, and \emph{Ours-MLP} denotes the method that uses an MLP as the decoder. We trained all the methods with the same data partition and in the same setup for a fair comparison. The methods used in our comparison are:

\begin{itemize}
\item Fully Convolutional Autoencoder (FCAE): We trained an FCAE with the same encoder used in our methods and a decoder resembling our vanilla decoder. The main difference with our method is that the FCAE predicts the complete point cloud (2048 points) rather than the missing part. EMD is used as a loss function.
\item Morphing and Sampling Network(MSN) \cite{sa:Liu2020}: We trained the MSN method to predict a coarse complete point cloud (2048 points). This coarse point cloud is further enhanced using a residual network. EMD is used as a loss function.
\item Point Fractal Network (PF-Net) \cite{sa:Huang2020}: PF-Net predicts the missing part rather than the overall shape of the point cloud. To do so, it uses a multi-stage completion loss and adversarial loss, whose input is the incomplete point cloud (2048 points), and outputs are the missing point cloud (1024 points). To compare against the other methods, we merged the input and the predicted point clouds and sampled 2048 points using farthest point sampling. The multi-stage completion loss is equal to the CD of the point clouds for each stage.
\end{itemize}

To compare our proposal with other state-of-the-art methods, we use the evaluation metric used in \cite{sa:Huang2020}. This metric is two-fold: $Pred \rightarrow GT$ error and $GT \rightarrow Pred$ error. The former error computes the average squared distance from each point in the predicted cloud to the ground-truth's closest point. This error measures how different the prediction is from the ground truth. The later error computes the average square distance from each point in the ground truth to the predicted cloud's closest point. This error measures how different the ground-truth is from the prediction. Note that the sum of errors $(Pred \rightarrow GT) + (GT \rightarrow Pred)$ is the Chamfer distance.

\begin{figure}[t]
\centering
\includegraphics[width=\columnwidth]{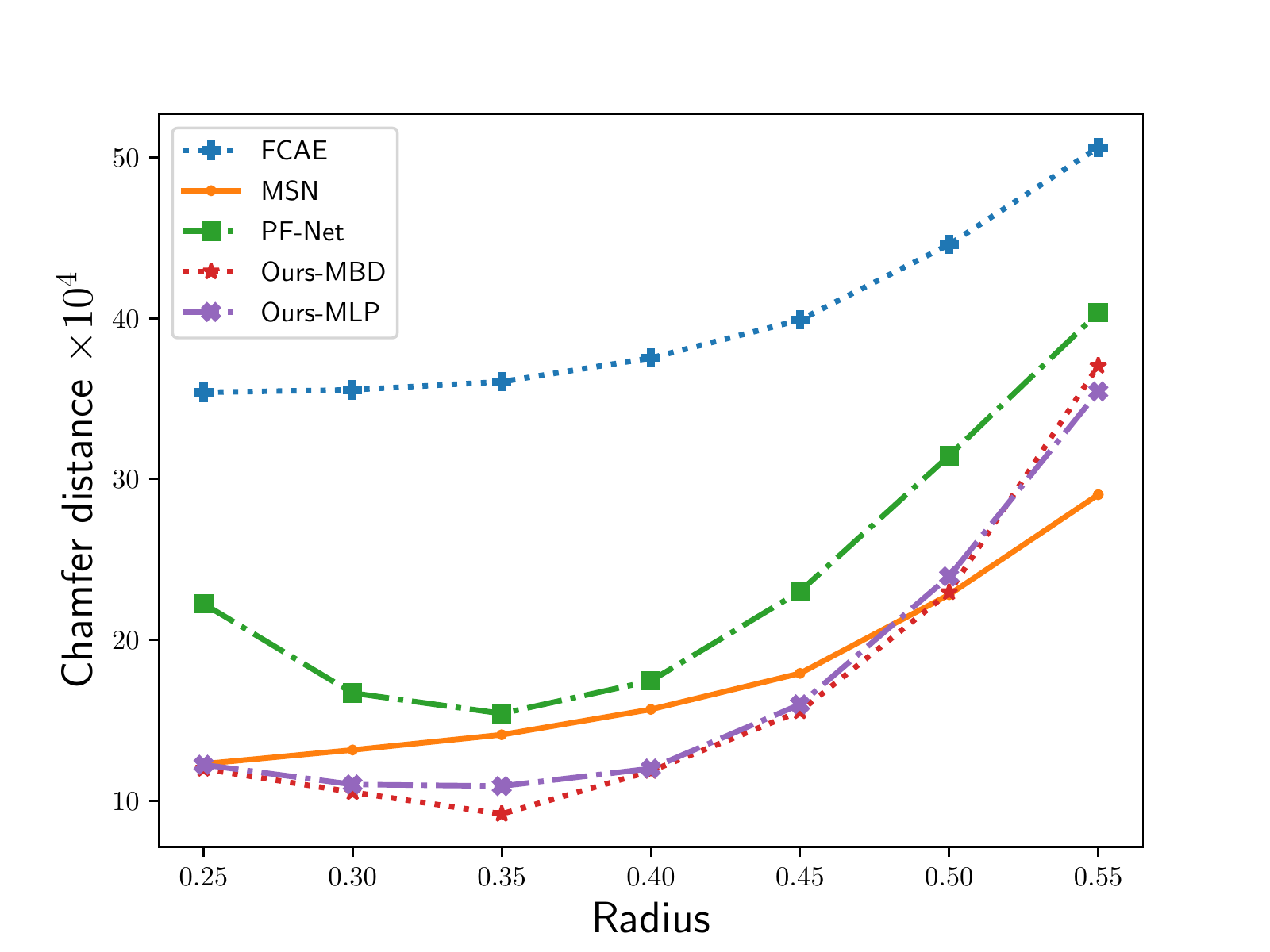}
\caption{Robustness against variable size of missing parts. Our methods are robust when the size of missing part is similar to the size of the trained missing part.}
\label{fig:robustness}
\end{figure}

The results are presented in Table \ref{tab:comparisonShapeNet}. \emph{Ours-MBD} outperforms all the methods in terms of the $Pred \rightarrow GT$ error. Ours-MBD computes a point cloud that is similar to the ground truth, in terms of the average distances of points in the predicted shape. Also, Ours-MLP outperforms all the methods in terms of the $GT \rightarrow Pred$ error. The difference is slight concerning the same error in Ours-MBD so that if we sum both errors (to obtain the Chamfer distance), Ours-MBD is still the best algorithm in the comparison. Our method performs well even for the harder classes in the dataset, namely Bag, Cap, and Lamp. The problem with classes Bag and Cap is the unbalanced number of shapes in the dataset (Bag contains 54 training shapes and Cap contains 39 training shapes), and the problem with Lamp is the high intra-class variability. We believe that these factors prevent automatic shape completion algorithms from generalizing and performing well.
Nevertheless, our method gets a notorious improvement concerning other techniques in such classes. Our method's advantage is the combination of missing part prediction with the refinement: in hard examples, the missing part prediction computes a coarse output, which is subsequently corrected by the refinement. In Section~\ref{Sec:Ablation}, we present an ablation study with quantitative results about the importance of the refinement.

Figure \ref{fig:comparison} shows qualitative results. Our methods (Ours-MBD and Ours-MLP) compute better completions for the shown examples. By definition, our methods preserve the original input geometry while focusing on the computation and refinement of the missing part. FCAE produces a fair coarse completion, but many of the details of the shape are lost. Similarly, MSN~\cite{sa:Liu2020} focuses on predicting the entire shape, but it still struggles to reproduce the geometric details of the shape. In contrast, PF-Net~\cite{sa:Huang2020} directly computes the missing part; however, there is no good integration between the partial input and the predicted missing part. Our method's main result is the ability to produce good missing parts that are effectively integrated through the refinement.

\begin{table*}[t]
\centering
\selectfont\begin{tabular}{lll||ll}
\hline
{\bftab Categories} & {\bftab Ours-MBD} & {\bftab Ours-MBD w/o Ref} & {\bftab Ours-MLP} & {\bftab Ours-MLP w/o Ref} \\ \hline
Airplane & \textbf{2.147} / \textbf{2.528} & 2.541 / 3.096 & \textbf{2.892} / \textbf{2.212}	& 3.652 / 2.573 \\
Bag & \textbf{9.975} / \textbf{6.731} & 12.580 / 7.745	& \textbf{14.232} / \textbf{5.748} & 15.584 / 9.849 \\
Cap & \textbf{8.596} / \textbf{2.500} & 13.892 / 4.809	& \textbf{17.016} / \textbf{2.527} & 25.148 / 5.877 \\
Car & \textbf{6.539} / \textbf{2.927} & 6.875 / 3.388	& \textbf{8.683} / \textbf{2.516} & 9.582 / 2.989 \\
Chair & \textbf{4.371} / \textbf{2.259} & 4.863 / 3.104	& \textbf{6.049} / \textbf{2.051} & 6.659 / 3.021 \\
Guitar & \textbf{1.329} / \textbf{2.483} & 1.648 / 2.546 & \textbf{1.822} / 2.381 & 2.414 / \textbf{2.307} \\
Lamp & \textbf{12.712} / \textbf{10.545} & 16.308 / 12.081	& \textbf{15.675} / \textbf{9.389} & 25.197 / 12.805 \\
Laptop & \textbf{2.239} / \textbf{1.551} & 2.827 / 2.318	& \textbf{2.832} / \textbf{1.430} & 4.111 / 1.861 \\
Motorbike & \textbf{4.897} / \textbf{5.946} & 5.253 / 6.480	& \textbf{6.197} / \textbf{4.162} & 7.504 / 5.616 \\
Mug & \textbf{4.467} / \textbf{2.963} & 5.571 / 4.053	& \textbf{7.464} / \textbf{3.054} & 9.166 / 4.731 \\
Pistol & \textbf{4.137} / \textbf{4.539} & 4.917 / 5.413	& \textbf{5.399} / \textbf{3.952} & 7.285 / 4.837 \\
Skateboard & \textbf{2.250} / \textbf{2.295} & 2.256 / 3.121	& \textbf{2.693} / \textbf{1.994} & 3.115 / 3.401 \\
Table & \textbf{5.556} / \textbf{2.777} & 6.390 / 4.317	& \textbf{6.722} / \textbf{2.737} & 7.584 / 4.688 \\ \hline
{\bftab Average} & \textbf{5.324} / \textbf{3.849} & 6.609 / 4.806	& \textbf{7.514} / \textbf{3.396}	& 9.769 / 4.966
\\ \hline
\end{tabular}
\caption{Refinement is important to improve the quality of the completion. Our methods consistently outperform the results against not using refinement. The improvement is even more evident in challenging classes such as Cap and Lamp. Results are scaled by $10,000$.}
\label{tab:ablation}
\end{table*}

\subsection{Robustness Test}
In this experiment, we aim at evaluating the ability to perform the completion with a varying size of removed geometry. The models used in this experiment are the same trained models from the comparison in Section~\ref{Sec:Comparison}. Recall that each input shape undergoes a splitting operation where a partial shape and a missing part are computed during training. The missing part's location is random, and the size depends on a given radius around the selected point. In the previous comparison, we used a radius of 0.35. We measure the robustness of the trained models to perform the completion when the test shape has a missing part with different sizes. We vary the radius from 0.25 to 0.55 and compute the average Chamfer distance for every algorithm. Figure~\ref{fig:robustness} shows the results.

Our methods (Ours-MBD and Ours-MLP) consistently outperform the other methods when the missing part's size varies from 0.25 to 0.45. Nevertheless, the performance of our methods degrades when the radius is above 0.5. In our opinion, the reason for this behavior is the extent of the missing part that is not well covered by our fixed number of points to represent the missing part. In this case, methods that predict the complete point cloud show better robustness. One of our methods' main limitations is the lack of adaptativeness to represent the missing part according to its extent. The computation of adaptive varying-size point clouds via neural networks is a promising idea to incorporate in our approach in future researches to increase robustness.

\subsection{The Importance of Refinement}
\label{Sec:Ablation}

An essential contribution of our methods is the combination of missing part prediction plus the final refinement. In this experiment, we show how important is the refinement in the completion result. We take the networks trained in the experiment of Section~\ref{Sec:Comparison} and evaluate the completion performance with and without the refinement module. Table~\ref{tab:ablation} shows the effect of using the Point Refiner Network in our models.

\begin{figure}[t]
\centering
\includegraphics[width=0.8\columnwidth]{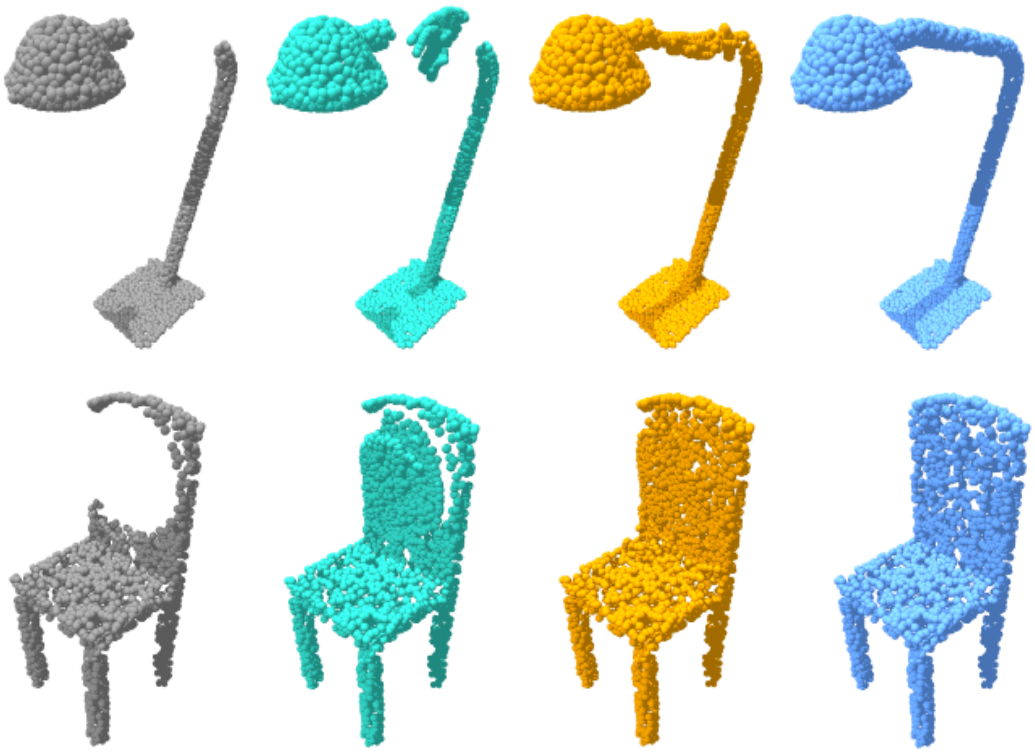}
\caption{The refinement is a key ingredient in our proposal. The predicted missing part is often a coarse representation which is considerably enhanced by the point refiner network. From left to right: input, input+predicted missing part, input+refined missing part, ground-truth. }
\label{fig:spc}
\end{figure}

The results show the importance of refinement in the final output. The improvement using the refinement is consistent in our two Ours-MBD and Ours-MLP and concerning both errors, $Pred \rightarrow GT$, and $GT \rightarrow Pred$. Note that the more significant gain is in the challenging classes: Bag, Cap, and Lamp. In effect, these results demonstrate that the refinement can be a useful complement to the missing part prediction, to the point that it can help deal with unbalanced data and high intra-class variability. Figure~\ref{fig:spc} shows two examples of the effect of the refinement in our proposal.

\section{Conclusions and future work}
\label{Sec:Conclusions}
We presented a novel architecture for point cloud completion that emphasizes the combination of missing part prediction and point cloud refinement to improve the completion. Unlike other architectures that predict the overall shape, our method retains the existing geometry and refined details while focusing on predicting and integrating the missing part. Our method also outperforms a similar method: PF-Net, which also aims at predicting the missing region. It is also worth noting that our architecture is robust, and it tends to reduce the error in challenging classes. As future work, we believe that an adaptive strategy to compute missing parts whose size is relative to the amount of geometry to represent would be useful in our approach.

{\small
\bibliographystyle{ieee_fullname}
\bibliography{egbibsample}
}

\end{document}